\documentclass[11pt,a4paper]{article}
\usepackage[hyperref]{acl2017}
\usepackage{times}
\usepackage{latexsym}
\usepackage{tikz, pgfplots}
\usepackage{framed}
\usepackage{url}
\usepackage{colortbl}
\usepackage{multirow}
\usepackage{enumitem}
\usepackage{booktabs}
\usepackage [english]{babel}
\setlist{nolistsep}
\usepackage{fancyhdr} 
\usepackage{color}
\usepackage{colortbl}
\usepackage{graphicx}
\newcommand{\ra}[1]{\renewcommand{\arraystretch}{#1}}
\newcolumntype{P}[1]{>{\centering\arraybackslash}p{#1}}
\pgfplotsset{compat=1.12}

\newcommand\mk\textsc
\pgfplotsset{compat=1.12} 
\definecolor{Gray}{gray}{0.9}
\definecolor{LightGray}{gray}{0.91}

\aclfinalcopy 

\title{Learning Lexico-Functional Patterns for First-Person Affect}

   \author{
   \textbf{Lena Reed, Jiaqi Wu, Shereen Oraby} \\
   \textbf{Pranav Anand and Marilyn Walker} \\
    University of California Santa Cruz \\ {\tt \{lireed,jwu64,soraby,panand,mawalker\}@ucsc.edu} \\
   }

\date{}

\begin{document}
\maketitle
\begin{abstract}
	Informal first-person narratives are a unique resource for
        computational models of everyday events and people's affective
        reactions to them.  People blogging about their day tend not
        to explicitly say {\it I am happy}. Instead they describe
        situations from which other humans can readily infer their
        affective reactions. However current sentiment dictionaries
        are missing much of the information needed to make similar
        inferences.  We build on recent work that models affect in
        terms of lexical predicate functions and affect on the
        predicate's arguments. We present a method to learn proxies
        for these functions from first-person narratives. We construct
        a novel fine-grained test set, and show that the patterns we
        learn improve our ability to predict first-person affective
        reactions to everyday events, from a Stanford sentiment
        baseline of .67F to .75F.
\end{abstract}

\section{Introduction}
\label{intro-sec}

Across social media, thousands of posts daily take the form of
informal \textsc{first-person narratives}. These narratives provide a
rich resource for computational modeling of how people feel about the
events they report on. Being able to reliably predict the affect a
person may feel towards events they encounter has a range of potential
applications, including monitoring mood and mental health
\cite{Isaacsetal13} and getting conversational assistants to respond
appropriately \cite{Bowdenetal17}.  Moreover, as these narratives
are told from the perspective of a protagonist, this research could be
used to understand other types of protagonist-framed narratives, like
those in fiction.

We are interested in the opinions that a protagonist has, not the 
author per se. This is sometimes referred to as internal sentiment 
or self reflective sentiment. While in many situations that is overlaid 
with the author's opinions, in first-personal narratives, because the 
author is the protagonist, the two perspectives align. Here, we use 
the term {\it affect} to reference this protagonist-centered notion of opinion.

A central obstacle to reliable affect prediction is that that people
tend not to {\em explicitly} flag their affective state, by saying 
{\it I am happy}.  Large-scale sentiment dictionaries focus
on compiling lexical items that bear a consistent affect all on their
own \cite{Wilsonetal05}. But people tend to describe situations, such as {\it My friend
  bought me flowers}, or {\it I got a
  parking ticket}, from which other humans can readily infer their
{\em implicit} affective reactions.

One approach to this problem aims to directly learn units larger than
a lexical item that
reliably bear some marker of polarity or emotion
\cite{Vuetal14,Lietal14,DingRiloff16,Goyaletal10,Russoetal15,Kiritchenkoetal14,Reckmanetal13}.

\begin{table}[ht]
\begin{center}
\begin{small}
\begin{tabular}{cccc|c}
\toprule
$X$ & $Y$ & $E_{hve}$ & $E_{lck}$ & Example \\
\cmidrule(rl){1-2} \cmidrule(rl){3-5} 
+ & + & + & -  & {\it I have a new kitten.}\\ 
+ & - & - & +  & {\it I got a parking ticket.} \\
- & + & - & + & {\it My rival got a prize.} \\
- & - & + & -  & {\it My rival got a reprimand.} \\
\cmidrule(rl){1-5}
 \multicolumn{5}{l}{\emph{X} have/lack \emph{Y}}\\
\bottomrule
	\end{tabular}
\end{small}
	\caption{Functions for verbs of possession. \label{func-have}}
\end{center}
\end{table}

Another approach aims to model the speaker's affect to an event 
compositionally, e.g. \newcite{AnandReschke10} (A\&R) proposed that
the affect a lexical predicate communicates should be modeled as an n-ary
function, taking as inputs the affect that the speaker bears
towards each participant. Table~\ref{func-have} contains
A\&R's functions for verbs of possession: a state in which {\it X has
  Y} or {\it X lacks Y} does not convey a clear affect unless we
know what the speaker thinks of both $X$ and $Y$. If the speaker has
positive affect toward both $X$ and $Y$ (Row 1), then we infer
that her attitude toward the event is positive, but if either is
negative, then we infer that the speaker
is negative toward the event. Similarly, \newcite{Rashkinetal16} represent
the typical affect communicated by particular predicates via connotation frames. 
Here we are finding the internal sentiment of the speaker, or, as Rashkin et al.
refer to it, the "mental state" of the speaker.  

Inspired by A\&R's framework, our work learns lexico-functional
patterns (patterns involving lexical items or pairs of lexical items
in specific grammatical relations that we show to capture
functor-argument relations in A\&R's sense), about the effects of
combining particular arguments with particular verbs (event types)
from first-person narratives.  Our novel observation is that learning
these compositional functions is greatly simplified in the case of
first-person affect. People bear positive affect to themselves, so
sentences with first-person elements, e.g. {\it I/we/me}, reduce the
problem for an approach like A\&R's to learning the polarity that
results from composing the verb with {\it only one} of its arguments,
i.e. only Rows 1, 2 in Table~\ref{func-have} need to be learned for
first person subjects. First-person narratives are full of such
sentences. See Table~\ref{pos-neg-sent-fig}. We show that the learned
patterns are often consonant with A\&R's predictions, but are richer,
including e.g. many private state descriptions
\cite{Wiebeetal04,Wiebe90}.

\begin{table}[t!h]
\centering
\begin{small}
\ra{1.3}
\begin{tabular}{|p{2.9in}|}
\hline
\textbf{Positive Sentences} \\
\hline \hline 
 We had a marvelous visit and drank coffee and ate homemade chocolate chip cookies. \\ 
\hline 
 Now, I could swim both froggy and free style swimming!!\\
\hline
  \hline  
 \textbf{Negative Sentences}   \\
\hline
 But last week, he said that he doesn't know if he has the same feelings for me anymore.\\
\hline
I didn't want to lose him.  \\
\hline
\end{tabular}
\caption{Sentences from the training data \label{pos-neg-sent-fig}}
\end{small}
\end{table}

In addition, we demonstrate that these lexico-functional patterns 
improve the performance of several off-the-shelf sentiment
analyzers. We show that Stanford sentiment
\cite{socheretal13} has a best performance of 0.67 macro
F on our test set. We then supplement it with our
learned patterns and demonstrate significant improvements.
 Our final ensemble achieves 0.75 F on the test set. 
We discuss related work in more detail in
Sec.~\ref{rel-sec}.

\section{Bootstrapping a First-Person Sentiment Corpus}
\label{corpus-sec}

We start with a set of first-person narratives (weblogs) drawn from
the Spinn3r corpus, that cover a wide range of topics
\cite{Burtonetal09,GordonSwanson09}. To reduce noise, we restrict the
blogs to those from well-known blogging sites
\cite{DingRiloff16}, and select 15,466 stories whose
length ranges from 225 to 375 words.

 \begin{table}[t!bh]
\begin{scriptsize}
\centering
\begin{tabular}{@{}p{.87in}|p{.2in}|p{1.65in}@{}}\toprule
\bf  Pattern Template &\bf Class & \bf  Example Instantiations \\ \midrule
    $<$subj$>$ ActVP & neg  & {\bf $<$I$>$ cry} at the thought of it and I'm crying now. \\ \midrule
    $<$subj$>$ ActInfVP &pos & {\bf $<$I$>$ got to swim} from the boat to a little sandbar. \\ \midrule
    $<$subj$>$ AuxVP Dobj &neg& Yesterday {\bf $<$it$>$ was} my 1st {\bf molar}, today it's my 2nd molar. \\ \midrule
    \midrule
    ActVP $<$dobj$>$ &pos& As I bake often, I have {\bf used $<$several different kinds of recipes$>$}.  \\ \midrule
    PassInfVP $<$dobj$>$ &pos & When we arrived at the Embarcadero, we were {\bf surprised} to find {\bf $<$ a music festival taking place$>$}... \\ \midrule

        Subj AuxVP $<$dobj$>$  &neg & Our {\bf relationship was $<$non existant$>$} for over a year after that.  \\ 
    \midrule \midrule
    NP Prep $<$np$>$  &neg& he hurt me countless times but I still forgave him and i still tried to prove to him that I did {\bf care for $<$him$>$}.  \\ \midrule
    ActVP Prep $<$np$>$ &neg& I didn't think anything of it until I thought about when he {\bf cheated on $<$me$>$}.  \\ \midrule
    InfVP Prep $<$np$>$ &pos&...my friend from college who was so generous {\bf to offer} his place to {\bf $<$us$>$}...  \\ \midrule
 \bottomrule
 \end{tabular}
\end{scriptsize}
\vspace{-.1in}
  \caption{\label{pattern-types} AutoSlog-TS Templates and Example Instantiations}
\end{table}

We hand-annotate a set of 477 positive and 440 negative stories,
and use these to bootstrap a larger set of 1,420 negative and 2,288
positive stories. To bootstrap, we apply AutoSlog-TS, a weakly
supervised pattern learner that only requires training sets os stories
labeled broadly as {\sc positive} or {\sc negative}
\cite{Riloffetal96,RiloffWiebe03}. AutoSlog uses a set of syntactic
templates to define different types of linguistic expressions.  The
left-hand side of Table~\ref{pattern-types} lists examples of AutoSlog
patterns and the right-hand side illustrates a specific
lexical-syntactic pattern that corresponds to each general pattern
template, as instantiated in first-person stories.\footnote{The
  examples are shown as general expressions for readability, but the
  actual patterns must match the syntactic constraints associated with
  the pattern template.} When bootstrapping a larger positive and
negative story corpus, we use the whole story, 
not just the first person sentences.

The left-hand-side of Table \ref{pattern-types} shows that the learned
patterns can involve syntactic arguments of the verbal predicate,
which means that these patterns are proxies for one column of verbal
function tables like those in Table \ref{func-have}.  However, they
can also include verb-particle constructions, such as {\it cheated
  on}, or verb-head-of-preposition constructions. In each case though,
because these patterns are localized to a verb and only one element,
they allow us to learn highly specific patterns that could be
incorporated into a dictionary such as +-Effect
\cite{ChoiWiebe14}. AutoSlog simultaneously harvests both
(syntactically constrained) MWE patterns and more compositionally
regular verb-argument groups at the same time.

AutoSlog-TS computes statistics on the strength of association of each
pattern with each class, i.e. P({\sc positive} $\mid$ $p$) and P({\sc
  negative} $\mid$ $p$), along with the pattern's overall
frequency. We define three parameters for each class: $\theta_f$, the
frequency with which a pattern occurs, $\theta_p$, the probability
with which a pattern is associated with the given class and
$\theta_n$, the number of patterns that must occur in the text for it
to be labeled. These parameters are tuned on the dev set
\cite{Riloffetal96,Orabyetal15,RiloffWiebe03}.

To bootstrap a larger corpus, we want settings that have lower recall but very
high precision. We  select $\theta_p = 0.7$, 
$\theta_f = 10$ and $\theta_n = 3$ for the positive class and $\theta_p = 0.85$, 
$\theta_f = 10$ and $\theta_n = 4$ for the negative class for bootstrapping.

\section{Experimental Setup}

Our experimental setup involves first creating a corpus of
training and test {\bf sentences}, then applying AutoSlog-TS 
a second time to learn linguistic patterns. We then set up
methods for cascading classifiers to explore whether ensemble classifiers
improve our results. 
%

\noindent{\bf Training Set:}  From the bootstrapped set of stories, 
we create a corpus of sentences. A critical simplifying assumption of our
method is that a multi-sentence story can be labelled as a whole as
positive or negative, and that each of its sentences {\bf inherit}
this polarity. This means
we can learn the polarity of events in such narratives from their
(noisy) inherited polarity without labelling individual sentences. 
Our training set consists of 46,255 positive and 25,069 negative sentences. 

\noindent{\bf Test Set:}  We create the test set
by selecting 4k random first-person sentences.  
First-person sentences either contain an explicit first person marker, i.e. {\it we} and 
{\it my} or start with either a progressive verb or pleonastic {\it it}. 
To collect gold labels, we
designed a qualifier and a HIT for Mechanical Turk, and put these out
for annotation by 5 Turkers, who label each instance as positive,
negative, or neutral.  To ensure the high quality of the test set, we
select sentences that were labelled consistently positive or negative
by 4 or 5 Turkers. We collected 1,266 positive and 1,440 negative sentences.

\noindent{\bf Dev Set:} We created the dev set using the same 
method as the test set, having Turkers annotate 2k random first-person 
sentences. We collected 498 positive and 754 negative sentences. The 4k 
test and dev sentences available for download at https://nlds.soe.ucsc.edu/first-person-sentiment.

\noindent{\bf AutoSlog First-Person Sentence Classifier.}  In order to
learn new affect functions, we develop a second {\bf sentence-level}
classifier using AutoSlog-TS.  We run AutoSlog over the training
corpus, using the dev set to tune the parameters $\theta_f$,
$\theta_p$ and $\theta_n$ \cite{Riloffetal96}, in order 
to maximize macro F-score.  Our best parameters on
the dev set for positive is $\theta_f$=18, $\theta_p$=0.85 and $\theta_n$=1 
and for negative is $\theta_f$=1, $\theta_p$=0.5 and $\theta_n$=1. We specify that if
the sentence is in both classes we rename it
as neutral.  We will refer to this classifier as the { AutoSlog
  classifier}. 
\vspace{.025in}

\noindent{\bf Baseline First-Person Sentence Classifiers.}  Our goal
is to see whether the knowledge we learn using AutoSlog-TS complements
existing sentiment classifiers.  We thus experiment with a number of
baseline classifiers: the default SVM classifier from Weka with
unigram features \cite{WittenFrank05}, a version of the NRC-Canada
  sentiment classifier \cite{MohammadKZ2013}, 
provided to us by \newcite{QadirRiloff14}, and the Stanford 
Sentiment classifier \cite{socheretal13}.

\noindent{\bf Retrained Stanford.} The Stanford Sentiment classifier  is a 
based on Recursive Neural Networks, and trained 
 on a compositional Sentiment Treebank, which includes fine-grained sentiment 
 labels for 215,154 phrases from 11,855 sentences from movie reviews. It can 
 accurately predict some compositional semantic effects and handle negation. 
 However  since it was trained on movie reviews, it is 
 likely to be missing labelled data for some common phrases in our blogs. Thus 
 we also retrained it ({\sc retrained stanford}) on high precision 
 phrases from AutoSlog extracted from our training data of positive and negative 
 blogs. This provides  67,710 additional phrases, including 58,972 positive phrases 
 and 8,738 negative phrases. The retrained model includes {\bf both} the labels 
 from the original Sentiment Treebank and the AutoSlog high precision phrases.

\section{Results and Analysis}
\label{results-sec}
We present our experimental results and analyze
the results in terms of the lexico-functional linguistic
patterns we learn. 
\vspace{.025in}

\noindent{\bf Baseline Classifiers.} 
Rows 1-3 of Table~\ref{table:results} show the results for the three
baselines, in terms of F-score for each class and the macro F. 
Stanford outperforms both NRC and SVM, but misses many cases of positive
sentiment. 
\vspace{.025in}

\noindent{\bf AutoSlog Classifier.} Row 4 of
Table~\ref{table:results} shows the results for the AutoSlog classifier. 
Although AutoSlog itself does not perform
highly, the patterns that it learns represent a
different type of knowledge than what is contained in many sentiment
analysis tools.  We therefore hypothesized that a cascading classifier,
which supplements one of the baseline sentiment classifiers with
the lexico-functional patterns that AutoSlog learns
might yield higher performance. 
\vspace{.025in}

\noindent{\bf Retrained Stanford.} Row 5 of
Table~\ref{table:results} shows the results for {\sc retrained stanford}. 
The F-scores for {\sc retrained stanford} are almost identical to
the standard Stanford classifier. This may be  because our data is
a small percentage of the entire number of phrases used in
training Stanford.  Although
{\sc retrained stanford} prioritizes our phrases, 
it would not make sense to  remove the original training data.

\vspace{.025in}

\noindent{\bf Cascading Classifiers.}  We implement cascading
classifiers to test our hypothesis.  The cascade classifier has
primary and secondary classifiers, and we invoke the secondary
classifiers only if the primary assigns a prediction of {\it neutral}
to a test instance, which reflects the lack of sentiment-bearing
lexical items.  We also have a cascade classifier with a tertiary classifier,
which is invoked in the same fashion as the secondary classifier after
the primary and secondary classifiers have been run. The cascading classifiers
are named in the order the classifier is employed, primary, secondary or 
primary, secondary, tertiary. For our cascading classifiers, we combine our 
baseline classifiers (NRC and Stanford), with our
AutoSlog classifier. We do not use SVM as a primary classifier since it has 
no neutral label.  The results for the cascading experiments are
in Rows 6-9 of Table~\ref{table:results}.  

Cascading NRC and AutoSlog provides the best performance, 
improving both the positive and negative
classes, for a macro F of 0.71. This shows that the learned implicit polarity
information from AutoSlog improves NRC's performance.

Since our best two-classifier cascade comes from combining NRC and
AutoSlog, we also test a cascade that adds Stanford or SVM. We
achieve our best macro F of 0.75
for the combination with SVM. 

\begin{table}[t!h]
\centering
\begin{small}
\begin{tabular}{|c|p{1.17in}|ccc|}
\hline
& \bf Classifier  &\bf Pos  &\bf Neg  & \bf Macro \\ 
&                 & \bf F1         & \bf F1          & \bf F \\ \hline
1 & \bf SVM  & 0.66& 0.60  & 0.64 \\ 
2 & \bf NRC   & 0.58 & 0.69  & 0.64 \\ 
3 & \bf Stanford   & 0.54  &  0.73 & 0.67 \\ 
\hline  
4 & \bf AutoSlog (ASlog)   & 0.11 & 0.68 & 0.53 \\ 
5 & \bf Retrained Stanford & 0.53 & 0.73 & 0.67 \\
\hline
6 & \bf NRC, ASlog    & 0.60  & 0.78 & 0.71 \\ 
7 & \bf Stanford, ASlog   & 0.55  & 0.76 & 0.70 \\ 
\hline
8 & \bf NRC, ASlog, Stanford  & 0.64 & \bf 0.79 & 0.74 \\  
9 & \bf NRC, ASlog, SVM   & \bf 0.70 & 0.78 & \bf 0.75 \\  
\hline
\end{tabular}
\end{small}
\caption{Test Set Results \label{table:results} }
\end{table}

\noindent{\bf Analysis and Discussion.}  Here we discuss how the patterns we learned
from AutoSlog  can
supplement the knowledge encoded in current sentiment classifiers, and
in newly evolving sentiment resources
\cite{Goyaletal10,ChoiWiebe14,Balahuretal12,RuppenhoferBrandes15}.  

\begin{table}[h!]
\begin{small}
\centering
\begin{tabular}{lp{3.5cm}p{3.5cm}|}
\hline
\bf POS PATTERNS & \bf Basic Entailment \\
\hline
\texttt{HAVE\_FUN} & property \\
\texttt{HAVE\_PARTY} & possession \\
\texttt{HEADED\_FOR} & location \\
 \hline
\bf NEG PATTERNS & \bf Basic Entailment \\
\hline
\texttt{HAVE\_CANCER} & property \\
 \texttt{LOST} & possession \\
 \texttt{NOT\_COME\_HOME} & location \\
 \texttt{NOT\_GOING\_KILL} & existence \\
 \hline
\end{tabular}
\caption{Highly predictable AutoSlog extracted case frames and functional description}\label{table:pos-neg-aslog}
\end{small}
\vspace{-.1in}
\end{table}

Tables~\ref{table:pos-neg-aslog} and ~\ref{table:not-in-ar}
illustrate several learned lexico-functional patterns for
positive events used in the AutoSlog classifier.
The patterns shown in Table~\ref{table:pos-neg-aslog} are predicted by
A\&R's framework, some functions of which can be seen in Table \ref{func-have}.
 For example, we find a range of basic state
descriptions ({\small {\texttt have\_party}}, {\small {\texttt
  have\_cancer}}) whose basic entailment category is either {\it
  possessive} or {\it property} state. Since $E_{have}$ is positive
for a first-person subject only if the object is positive, and
negative if the object is negative, we predict that {\it
  parties} are good to possess and that {\it cancer} is a bad property to
have. In this way, we can recruit the existing function for {\it have}
to induce new positive or negative things to ``possess.'' In line with
A\&R's claims,  many events are identified with their final results:
{\it headed for} results in being at a desired location, while
{\it not coming home} results in something failing to be at a desired
location. We find it a welcome result that our semi-supervised methods
yield patterns that correspond to the A\&R classes, thus validating our
suspicion that first-person sentences furnish a simplifying test ground
for discovering functional patterns in the wild.

However, many patterns are not covered by A\&R's general classes, see
Table~\ref{table:not-in-ar}.  Looking first at verbs, one major
correlation is between positive classes and public events and negative
classes and private states.  Verbs extracted from the positive class
tend to be eventive and agentive describing more
dynamic activities and interactions, such as \texttt{played},
\texttt{swim}, \texttt{enjoyed}, and \texttt{danced}. Even many
positive {\it have} uses are light verbs describing an activity
such as {\it have lunch}.

\begin{table}[t!h]
\begin{small}
\centering
    \begin{tabular}{lp{4.6cm}p{5.8cm}|}
\hline
\bf Description & \bf POS PATTERNS\\  \hline
activities &  \texttt{HAVE\_DINNER}, \texttt{HAVE\_WEDDING}\\
success & \texttt{GOT\_SEE},  \texttt{WENT\_WELL}\\
planning & \texttt{HEADED\_FOR}, \texttt{SET\_UP}
\\
free time & \texttt{HAVE\_TIME}, \texttt{TIME\_WITH} \\
social bonding & \texttt{PICTURE\_OF}, \texttt{OLD\_FRIENDS}\\
\hline
\bf Description & \bf NEG PATTERNS\\
\hline 
activities & \texttt{HAVE\_X-RAY}, \texttt{GET\_EXAM}\\
knowledge & \texttt{REALIZE}, \texttt{NOT\_KNOW\_WHAT}\\
unmet desire & \texttt{WANTS},  \texttt{NEED\_MONEY}\\
social bonding& \texttt{NOT\_TRUST}\\
\hline
\end{tabular}
\caption{Highly predictable AutoSlog case frames outside A\&R's functional system}\label{table:not-in-ar}
\end{small}
\vspace{-.1in}
\end{table}

Verbs from the negative class are strikingly different.  They are very
often stative, where the author is the
experiencer (cognitive subject) of that private state.   
While this state vs. event distinction is not one existing computational models of
sentiment or affect discuss explicitly, it replicates a finding
that consistently emerges in clinical psychology, one that is
explicitly argued for in cognitive-behavioral accounts of the mood that particular
activities evoke
\cite{lewinsohn1985unpleasant,macphillamy1982pleasant,Russoetal15}.
In addition, Table~\ref{table:not-in-ar} reveals several novel result
state categories. The success, planning, and unmet desire frames are
all ultimately about goal-fulfillment (or lack thereof). While 
the success and unmet desire cases could be understood as having
or lacking something, the planning cases indicate steps achieved
toward a desired end-state. Previous work
on learning affect from eventuality descriptions has largely focused on 
actions. Our results indicate that private state descriptions are another rich
source of evidence.

\section{Related Work}
\label{rel-sec}

Previous work learns phrasal markers of implicit polarity via
bootstrapping from large-scale text sources, e.g. \newcite{Vuetal14}
learn emotion-specific event types by extracting {\tt emotion,event}
pairs on
Twitter.  \newcite{Lietal14} uses Twitter to bootstrap `major life
events' and typical replies to those events.

\newcite{DingRiloff16} extract subj-verb-obj triples from blog posts. 
They then apply label propagation to spread
polarity from sentences to events.  
However, the triples they
learn do not focus on first-person experiencers.
They also filter private states out of
the verbs  used to learn their triples, whereas we have found
that verbs relating to private states such as {\it need, want} and
{\it realize} are important indicators of first-person affect.

\newcite{Balahuretal12} use the narratives produced by
the ISEAR questionnaire \cite{SchererWallbott86} for first-person
examples of particular emotions and extract sequences of subject-verb-object triples,
which they annotate for basic emotions.

Recent work has built on this idea, and developed methods to
automatically expand Anand \& Reschke's verb classes to create completely new lexical 
resources
\cite{Balahuretal12,ChoiWiebe14,Dengetal13,DengWiebe14,RuppenhoferBrandes15}.
Choi \& Wiebe's work comes closest to ours in trying to induce (not
annotate) lexical functions, but we attempt to infer these from
stories directly, whereas they use a structured lexical resource.



\section{Conclusion}
\label{conc-sec}

We show that we can learn lexico-functional linguistic patterns that
reliably predict first-person affect.  We constructed a dataset of
positive and negative first-person experiencer sentences and used them
to learn such patterns. We then showed that the performance of current
sentiment classifiers can be enhanced by augmenting them with these
patterns. By adding our AutoSlog classifier's results to existing
classifiers we were able to improve from a baseline 0.67 to 0.75 Macro
F with a cascading classifier of NRC, AutoSlog and SVM.  In addition,
we analyze the linguistic functions that indicate positivity and
negativity for the first person experiencer, and show that they are
very different. In first-person descriptions, positivity is often
signaled by active participations in events, while negativity involves
private states.  In future work, we plan to explore the integration of
these observations into sentiment resources such as the +-Effect
lexicon \cite{ChoiWiebe14}. We plan to apply these high precision
first-person lexical patterns beyond blog data and with
other person-marking.
\bibliography{nl}
\bibliographystyle{acl_natbib}

\end{document}